\documentclass[letterpaper]{article}
\usepackage{aaai}\nocopyright
\usepackage{times}
\usepackage{helvet}
\usepackage{courier}
\usepackage{graphicx}
\usepackage{subfig}
\usepackage{amsmath,amssymb,bm,color,url}

\usepackage{pgfplots}
\pgfplotsset{compat=1.9}
\pgfplotsset{every tick label/.append style={font=\scriptsize}}
\pgfplotsset{every label/.append style={font=\scriptsize}}

\newcommand{\citetext}[1]{\citeauthor{#1} \shortcite{#1}}

\makeatletter
\renewcommand\@biblabel[1]{}
\makeatother

\begin{document}

\title{Knowledge-Based Distant Regularization in Learning Probabilistic Models}

\author{
    Naoya Takeishi$^{\S,}$\thanks{A substantial portion of this work was performed while the authors were at the University of Tokyo.} \and Kosuke Akimoto$^{\dagger,}\footnotemark[1]$\\
    $^\S$RIKEN Center for Advanced Intelligence Project\\
    $^\dagger$Security Research Laboratories, NEC Corporation\\
    \texttt{naoya.takeishi@riken.jp}\hspace{2em}
    \texttt{k-akimoto@ab.jp.nec.com}
}

\maketitle


\begin{abstract}
Exploiting the appropriate inductive bias based on the knowledge of data is essential for achieving good performance in statistical machine learning. In practice, however, the domain knowledge of interest often provides information on the relationship of data attributes only distantly, which hinders direct utilization of such domain knowledge in popular regularization methods. In this paper, we propose the \emph{knowledge-based distant regularization} framework, in which we utilize the distant information encoded in a knowledge graph for regularization of probabilistic model estimation. In particular, we propose to impose prior distributions on model parameters specified by knowledge graph embeddings. As an instance of the proposed framework, we present the factor analysis model with the knowledge-based distant regularization. We show the results of preliminary experiments on the improvement of the generalization capability of such model.
\end{abstract}


\section{Introduction}
\label{intro}

The data-driven nature of statistical machine learning is one of the principal reasons for its success, whereas imposing an appropriate inductive bias is indispensable for gaining generalization capability.
Regularization is a prevailing methodology for imposing the inductive bias. Several types of popular regularization methods are known, such as Tikhonov regularization and sparsity regularization, and they have been commonly utilized in a wide range of tasks with strong theoretical supports. However, it is not always possible for those general regularization strategies to directly leverage the domain knowledge available on a given application since they are often built upon the statistical ``meta-knowledge,'' rather than the domain knowledge.

\begin{figure}[t]
    \centering
    \includegraphics[width=\linewidth]{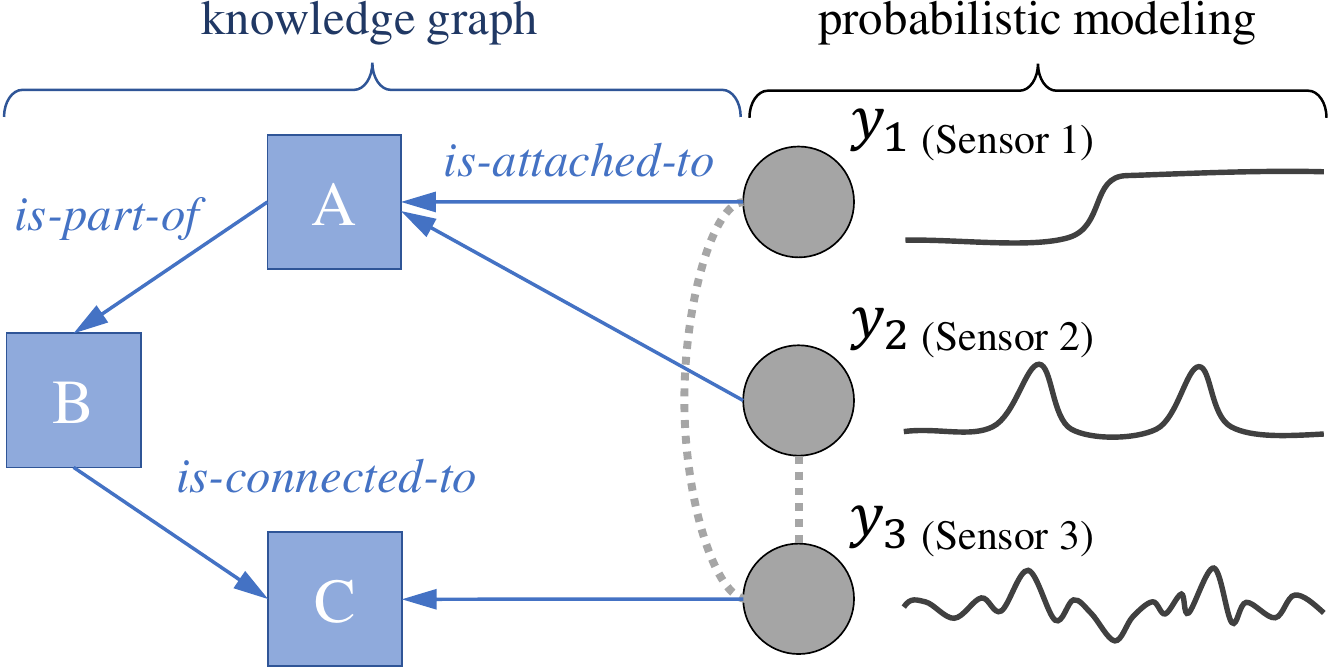}
    \caption{Schematic diagram of our motivating example, the analysis of sensor data of a plant. In such applications, even if the physical and/or statistical relations (dotted lines) between attributes, i.e., sensor readings, cannot be specified directly, somehow distant information on plant's instruments (blue arrows) are often available beforehand. We suppose such information can be encoded in a form of knowledge graphs. We would like to leverage such distant information to regularize learning of probabilistic models.}
    \label{fig:example}
\end{figure}

The usability of domain knowledge is one of the primal concerns in practices of machine learning.
If one could clearly specify the relationship of attributes and objects in data, it could then be utilized by the methods like structured sparsity regularization \cite{Huang11} and graph Laplacian regularization (see, e.g., \cite{Shahid16} and references therein).
Unfortunately, however, the domain knowledge of interest does not always provide direct information on the relationship of the attributes and the objects; instead, it often gives \emph{distant} information on them.
Our motivating example, depicted in Figure~\ref{fig:example}, is in the analysis of sensor data obtained from a plant.
Although the relations between plant's instruments and subsystems (e.g., how they are connected each other) can often be written down easily, the relations between the attributes (e.g., sensor IDs) can hardly be derived directly from them because to do so, we need to identify the mechanism and the physics of the instruments and the sensors, which is costly in many practices.

In this paper, we suggest a method to utilize such distant domain knowledge to regularize the estimation of statistical models.
Specifically, we present a framework of \emph{knowledge-based distant regularization} in learning probabilistic models for continuous-valued data such as sensor readings.
We aim to build a regularization principle using a knowledge graph, which is a multi-relational directed graph in which relations between entities are enumerated. This is because a knowledge graph is often useful for summarizing the domain knowledge of interest.
Let us revisit the example of sensor data analysis in Figure~\ref{fig:example}; engineers of a plant comprehend the relations between the instruments, such as \emph{is-part-of} and \emph{is-connected-to}, as well as the relations between the instruments and the sensors, i.e., \emph{is-attached-to}, according to which one can build a knowledge graph.
Note that what is specified on the data attributes is only the simple \emph{is-attached-to} information, and no direct relations between them are provided in general.

The central idea of the proposed method is to define prior distributions of model parameters using knowledge graph embeddings (see the excellent survey papers such as \cite{Nickel15,Wang17,Cai18}), which are continuous-valued representations of entities (and relations) in knowledge graphs.
The proposed method is flexible enough to utilize many types of knowledge graph embeddings, which form an active field of research recently. Hence, the development in the field can be immediately imported to the proposed framework.

The chief advantage of the proposed method, the distant regularization by knowledge graphs, is that it can be applied to a broad range of probabilistic models. We introduce the knowledge-based factor analysis model in this paper as an instance, but similar modification of existing probabilistic models is possible whenever the domain knowledge of interest is described in a form of knowledge graphs.


\section{Related Work}
\label{related}

The inductive bias is the essential element of statistical machine learning and usually introduced by designing features, hypothesis spaces, loss functions, regularization terms, and so on.
For example, the model family is usually selected according to user's knowledge of the data to be analyzed. If the user is sure about the relationship of attributes and objects in a dataset, a model tailored for the dataset (e.g., Bayesian networks) can be developed. This is difficult or impossible, however, when the knowledge is given in a distant form. In fact, there are a lot of researches on utilizing knowledge in various forms.

In what follows, we introduce only a part of the rich literature on knowledge utilization for machine learning and some other related topics that would be useful for considering the future direction of this work.

\subsubsection{Incorporating Knowledge as Graphs}

Actually, it has been actively studied how to utilize the prior knowledge that can be encoded in a knowledge graph.
For example, \citetext{Yao17} proposed an extension of latent Dirichlet allocation that models word tokens as well as entities in document corpora, where the entities are corroborated by a knowledge graph and have the corresponding embeddings. Also in the context of topic models, \citetext{Yao15} and \citetext{Hu16} suggested to utilize a taxonomy of entities in documents; a taxonomy, i.e., a set of hierarchical relations between entities, can be described as a (single-relational) graph.
Note that these studies mainly treat discrete-valued data such as document corpora.

Regularization based on the graph information is a major field of research also in other contexts.
For example, the graph sparsity regularization \cite{Jacob09,Huang11,Mairal13} imposes structured sparsity on parameters according to a graph that encodes prior knowledge on data attributes.
However, this method is not always applicable to utilizing the prior domain knowledge because in this method, all the data attributes must be identified in the graph and extension to multi-relational cases is not straightforward.

Another example of graph-based regularization is the methods using the graph Laplacian (see, e.g., \cite{Shang12,Pei15,Shahid16}), in which the Laplacian matrix of graphs that encode relations between objects and/or attributes is used for specifying the regularization term. However, such methods are also not always applicable in practice because we may not be able to compute a reliable Laplacian matrix due to the lack of precise knowledge of the relations between the data attributes and objects.

\subsubsection{Incorporating Knowledge as Constraints / Rules}

Researchers have been studying on methods to incorporate the knowledge described in a form of constraints or logical rules to well-known machine learning methods. To name a few, \citetext{Towell94} proposed the knowledge-based neural networks, \citetext{Fung03} suggested the knowledge-based support vector machines, and \citetext{Varol12} formulated the constrained latent variable model.
Moreover, there are studies on incorporating logical rules to existing statistical models. \citetext{Schiegg12} proposed the mixtures of Gaussian process regressors controlled by logical rules formulated by the Markov logic network \cite{Richardson06}.
And, several researchers proposed the topic models with logic \cite{Andrzejewski11,Mei14,Foulds15}, which are useful for modeling document corpora with prior knowledge of words.

\subsubsection{Statistical Relational Learning}

More generally, learning on relational data, which is often termed statistical relational learning (SRL), is an active area of research with a broad range of methods and applications (see, e.g., \cite{DeRaedtBook}). We do not describe the details here because of the vast amount of the literature in the field, but the methodology of SRL would be a promising tool in the future direction of this work.

\subsubsection{Distant / Weak Supervision by Knowledge}

Particularly in the context of supervised learning, there is a line of studies on utilizing user's heuristics or knowledge bases as weak supervision sources (see \cite{Ratner16,Varma17,Bach17} and references therein) with growing interests. In the settings of those studies, the knowledge should provide information on labels to be predicted.


\subsubsection{Posterior Regularization}

\citetext{Zhu14} proposed an elegant and flexible framework, termed RegBayes, for regularizing the posterior distribution of Bayesian inference. It can be utilized for efficiently incorporating one's knowledge in Bayesian inference, and in fact, \citetext{Mei14} utilized it for incorporating logical rules to a topic model.
Note that, while RegBayes tries to change the posterior, the method proposed in this work tries to change the prior, which may remind one of the empirical Bayes method.


\section{Background}
\label{back}

In this section, we first introduce the notations used in this paper to express the concepts regarding the probabilistic models.
Afterward, we give a brief explanation on knowledge graph embeddings for completeness, which readers familiar with them may skip.

\subsection{Probabilistic Models for Continuous-Valued Data}

The main subject of this study is the regularization in the estimation of probabilistic models for continuous-valued data. Formally, we denote a continuous-valued dataset by a set $\{\bm{y}_j \in \mathbb{R}^m \mid j=1,\dots,n\}$, where $\bm{y}_j$ is the observation with $m$ attributes (i.e., features) corresponding to the $j$-th object. And, $n$ denotes the number of objects (individuals, timestamps, etc.) in a dataset.
For instance, we may have readings of $m$ types of sensors at $n$ timestamps in the sensor data.
Note that each $\bm{y}_j$ may or may not be independent and identically distributed.
The probabilistic models for such data define probability density $p(\bm{y}_1,\dots,\bm{y}_n \mid \theta)$, where $\theta$ is a set of model parameters. Moreover, some models may have an additional set of random variables, denoted by $\{\bm{x}_j \in \mathbb{R}^{d_x}\}$, which is often termed latent variables.

We further distinguish the model parameters as $\theta=\{\bm{w}_1,\dots,\bm{w}_m,\bm\pi\}$; on one hand, let $\bm{w}_i\in\mathbb{R}^{d_w}$ denote the \emph{local parameter}\footnote{The dimensionality of $\bm{w}_i$ may depend on $i$, but for simplicity, we assume the same dimensionality for all $i$.} corresponding to the $i$-th attribute.
On the other hand, let $\bm\pi$ be the vector of \emph{global parameters} that are independent of the difference of attributes.
Let us give an example; the observation model of probabilistic principal component analysis \cite{Tipping99} is
\begin{equation*}
    p(\bm{y}_j \mid \bm{x}_j, \theta) = \mathcal{N} \left( \bm{y}_j \mid \bm{W}\bm{x}_j + \bm\mu, \ \sigma^2\bm{I} \right),
\end{equation*}
where $\bm{W}\in\mathbb{R}^{m \times d_x}$ is the factor loading matrix and $d_x$ is the dimensionality of latent variable $\bm{x}$.
In this model, the $i$-th local parameter, $\bm{w}_i$, corresponds to the $i$-th row of $\bm{W}$ (and the $i$-th element of $\bm\mu$), and $\sigma^2$ is an element of the global parameters, $\bm\pi$.

The regularization in the estimation of probabilistic models is often executed by setting prior distributions on the parameters, and many other popular regularization methods can also be interpreted as imposing prior distributions. In this paper, we follow this strategy to achieve the knowledge-based distant regularization; we place the prior distributions that are parameterized by the embeddings of knowledge graphs. The details are described in the next section.

\subsection{Knowledge Graph Embedding}

A knowledge graph is a multi-relational directed graph whose vertices correspond to entities and edges correspond to relations between the entities. Such graph can be described as a set of tuples $\{(h,r,t)\}$, where $h$ and $t$ denote the head entity and the tail entity respectively, and $r$ denotes the relation between them.
There are various large-scale knowledge bases which can be regarded as knowledge graphs, such as Freebase and GeneOntology, which are collections of knowledge on general entities of the world or on domain-specific concepts.

Representation learning for knowledge graphs is an active area of research recently. We do not enumerate the works and just refer to the excellent survey papers on the field \cite{Nickel15,Wang17,Cai18}.
Here, we introduce the very basic concepts in this field, which are needed when describing the proposed method.

In a typical setting, given a set of tuples that exist in a knowledge graph (i.e., positive tuples), we learn real-valued vectors $\bm{e}\in\mathbb{R}^{d_e}$ corresponding to entities in the knowledge graph. Such vectors are called \emph{embeddings} of entities. The learning is done so that some score function
\begin{equation}
    \psi(\bm{e}_\mathrm{h}, \ \bm{e}_\mathrm{t}; \ M_\mathrm{r}): \mathbb{R}^{d_e} \times \mathbb{R}^{d_e} \to \mathbb{R}
\end{equation}
becomes large for the positive tuples and small for the negative tuples. Here, $\bm{e}_\mathrm{h}$ and $\bm{e}_\mathrm{t}$ denote the embeddings of entities $h$ and $t$, respectively, And, $\psi$ is the score function that has the relation-specific parameter, $M_\mathrm{r}$, with regard to relation $r$.
In the latter part of this paper, as an instance of the knowledge graph embedding, we use a simple method called DistMult \cite{YangB15}, whose score function is defined as follows:
\begin{equation}
    \psi(\bm{e}_\mathrm{h}, \ \bm{e}_\mathrm{t}; \ \bm{m}_\mathrm{r}) = \bm{e}_\mathrm{h}^\top \operatorname{diag}(\bm{m}_\mathrm{r}) \bm{e}_\mathrm{t},
    \label{eq:distmult}
\end{equation}
where $\bm{m}_\mathrm{r}\in\mathbb{R}^{d_e}$ is the relation-specific parameter.
There are several manners of optimization for learning $\bm{e}$ and $M$, such as energy-based optimization and rank optimization. In this paper, as described below, we simply treat $\psi$ as a part of a logistic regressor and maximize the binary classification likelihood via negative tuple sampling.

Most studies on the knowledge graph embedding consider only the sparse structure of a multi-relational graph. Recently, there are several studies on fusing additional information associated with knowledge graphs such as continuous-valued attributes of entities \cite{Zhang16}, literals \cite{Kristiadi18}, text descriptions \cite{Xie16,Fan17,Xiao17}, images \cite{OnoroRubio17}, and the multimodal information \cite{Pezeshkpour17,Thoma17}. Their primary concern is the improvement of the knowledge graph embeddings and their applications such as link prediction. The techniques utilized there would be of great interest also for our purpose, i.e., regularization in learning probabilistic models for continuous data.

\section{Proposed Method}
\label{proposed}

In this section, we describe the detail on the proposed method for \emph{distantly regularizing} the estimation of probabilistic models using knowledge graph embeddings.
First, the general framework, in which we do not premise special types of probabilistic models and knowledge graph embeddings, is presented. Afterward, as an instance of the framework, we introduce the factor analysis distantly regularized by a knowledge graph.

\subsection{General Framework}

Our purpose is to build a regularization principle for the estimation of the parameters of probabilistic models according to a knowledge graph that encodes (possibly distant) prior knowledge on the data-generating system. To this end, we propose to specify the prior distributions of the model parameters using knowledge graph embeddings.
Formally, consider a probabilistic model
\begin{equation}
    p(\bm{y}_1,\dots,\bm{y}_n \mid \theta),
    \label{eq:datalik}
\end{equation}  
where $\theta=\{\bm{w}_1,\dots,\bm{w}_m,\bm\pi\}$.
In the proposed framework, we specify the prior distributions on the (part of) local model parameters, $\bm{w}_1,\dots,\bm{w}_m$, as follows.

First, we assume that we have a knowledge graph, in which entities corresponding to the (part of) data attributes exist, and that such entities appear in a set of positive tuples at least once. In the sensor data example (Figure~\ref{fig:example}), this assumption means that there is an entity like ``Sensor $i$,'' and the knowledge graph has a tuple like
\begin{equation*}
    (h=\text{Sensor}~i, \ r=\text{is-attached-to}, \ t=\text{Instrument~A})
\end{equation*}
for at least a part of attributes $i=1,\dots,m'$ ($1 \leq m' \leq m$).
In the following, we denote the embedding of the entity that corresponds to the $i$-th attribute of data by $\bm{e}_i$.
Note that in the knowledge graph of our interest, there would also be many entities that do not directly correspond to the attributes, and not all the attributes appear in the knowledge graph.
Also note that, while here we consider only the entities corresponding to the attributes, similar discussion is straightforward even when we consider entities corresponding to the objects in data.

Given the knowledge graph of this kind, we specify a prior distribution on $\bm{w}_i$ parameterized by $\bm{e}_i$, for $i=1,\dots,m'$, as a regularizer of the model training.
For example, for real-valued parameters, a canonical choice is the normal distribution, i.e.,
\begin{equation}
    p(\bm{w}_i \mid \bm{e}_i) = \mathcal{N}(\bm{w}_i \mid \bm{c}_\xi(\bm{e}_i), \ \bm{V}_\xi(\bm{e}_i)),
    \label{eq:paramprior}
\end{equation}
where $\bm{c}_\xi:\mathbb{R}^{d_e}\to\mathbb{R}^{d_w}$, $\bm{V}_\xi:\mathbb{R}^{d_e}\to\mathbb{R}^{d_w \times d_w}$, and $\xi$ denotes the set of parameters in $\bm{c}$ and $\bm{V}$.
With the priors on $\bm{w}_1,\dots,\bm{w}_{m'}$ specified in this manner, we compute their maximum a posteriori (MAP) estimations. For the remaining attributes $\bm{w}_{m'+1},\dots,\bm{w}_m$, if any, we compute the maximum likelihood estimations or MAP estimations with priors defined otherwise.

For the above framework, the embeddings of knowledge graph's entities must be prepared in some ways.
A promising way to prepare good embeddings is to utilize the model pretrained on a large-scale knowledge graph. However, such pretrained embeddings are usually not available for user-defined knowledge graphs that describe the domain knowledge. Hence, in many practices, the embeddings have to be learned simultaneously with the parameters of the probabilistic model.
Now suppose we define a score function $\psi$ for the embedding learning. In the framework, we propose to maximize binary classification likelihood:
\begin{equation}
    p((h,r,t)=\text{true} \mid \bm{e}_\mathrm{h}, \ \bm{e}_\mathrm{t}, \ M_\mathrm{r}) = \Phi(\psi(\bm{e}_\mathrm{h}, \ \bm{e}_\mathrm{t}; \ M_\mathrm{r})),
    \label{eq:kglik}
\end{equation}
where $\Phi$ is the sigmoid function. Note that any $\psi$ can be used here so long as its range is the real.
By setting the objective of the embedding learning in this way, the final objective function simply becomes the addition of the logarithms of data likelihood Eq.~\eqref{eq:datalik}, parameter priors Eq.~\eqref{eq:paramprior}, and knowledge graph likelihood Eq.~\eqref{eq:kglik}.

\begin{figure*}[t]
    \centering
    \begin{minipage}[t]{0.75\linewidth}
        \centering
        \subfloat[]{\label{fig:random}
        \begin{minipage}[t]{0.41\textwidth}
            \vspace{0cm}
            \begin{tikzpicture}[]
            \tikzstyle{every node}=[font=\scriptsize]
            \begin{axis}[width=\textwidth, xlabel={proportion of KG tuples [\%]}, ylabel={Ave. test NLL of FA}, ymajorgrids=true, xmajorgrids=true]
            \addplot+[error bars/.cd, y dir=both, y explicit] table[y error index=2] {summary_rain_random_dim5_triplerates.txt};
            \end{axis}
            \end{tikzpicture}
        \end{minipage}
        \begin{minipage}[t]{0.04\textwidth}\hfill\end{minipage}
        \begin{minipage}[t]{0.41\textwidth}
            \vspace{0cm}
            \begin{tikzpicture}[]
            \tikzstyle{every node}=[font=\scriptsize]
            \begin{axis}[width=\textwidth, xlabel={proportion of training data [\%]}, ylabel={Ave. test NLL of FA}, ymajorgrids=true, xmajorgrids=true, legend pos=outer north east]
            \addplot+[error bars/.cd, y dir=both, y explicit] table[y error index=2] {summary_rain_random_dim5_trterates_triple100.txt};
            \addplot+[error bars/.cd, y dir=both, y explicit] table[y error index=2] {summary_rain_random_dim5_trterates_triple0.txt};
            \legend{w/ KG, w/o KG}
            \end{axis}
            \end{tikzpicture}
        \end{minipage}
        \begin{minipage}[t]{0.13\textwidth}\hfill\end{minipage}
        }
        \par
        \subfloat[]{\label{fig:shift}
        \begin{minipage}[t]{0.41\textwidth}
            \vspace{0cm}
            \begin{tikzpicture}[]
            \tikzstyle{every node}=[font=\scriptsize]
            \begin{axis}[width=\textwidth, xlabel={proportion of KG tuples [\%]}, ylabel={Ave. test NLL of FA}, ymajorgrids=true, xmajorgrids=true]
            \addplot+[error bars/.cd, y dir=both, y explicit] table[y error index=2] {summary_rain_dim5_triplerates.txt};
            \end{axis}
            \end{tikzpicture}
        \end{minipage}
        \begin{minipage}[t]{0.04\textwidth}\hfill\end{minipage}
        \begin{minipage}[t]{0.41\textwidth}
            \vspace{0cm}
            \begin{tikzpicture}[]
            \tikzstyle{every node}=[font=\scriptsize]
            \begin{axis}[width=\textwidth, xlabel={proportion of training data [\%]}, ylabel={Ave. test NLL of FA}, ymajorgrids=true, xmajorgrids=true, legend pos=outer north east]
            \addplot+[error bars/.cd, y dir=both, y explicit] table[y error index=2] {summary_rain_dim5_trterates_triple100.txt};
            \addplot+[error bars/.cd, y dir=both, y explicit] table[y error index=2] {summary_rain_dim5_trterates_triple0.txt};
            \legend{w/ KG, w/o KG}
            \end{axis}
            \end{tikzpicture}
        \end{minipage}
        \begin{minipage}[t]{0.13\textwidth}\hfill\end{minipage}
        }
    \end{minipage}
    \caption{Results under (a) the \texttt{Random} data-partition scenario and (b) the \texttt{Shift} data-partition scenario. In each panel, left plot shows average test NLLs along different proportions of knowledge graph tuples used by the proposed method, and the right plot shows average test NLLs along different amounts of data used for the training. In the right plots, the test NLLs of two cases are shown: 100\% of the knowledge graph tuples was used by the proposed method (\emph{w/ KG}) or no knowledge graph tuples were provided (\emph{w/o KG}). In every plot, the means and the standard deviations for 10 random trials are shown.}
    \label{fig:}
\end{figure*}
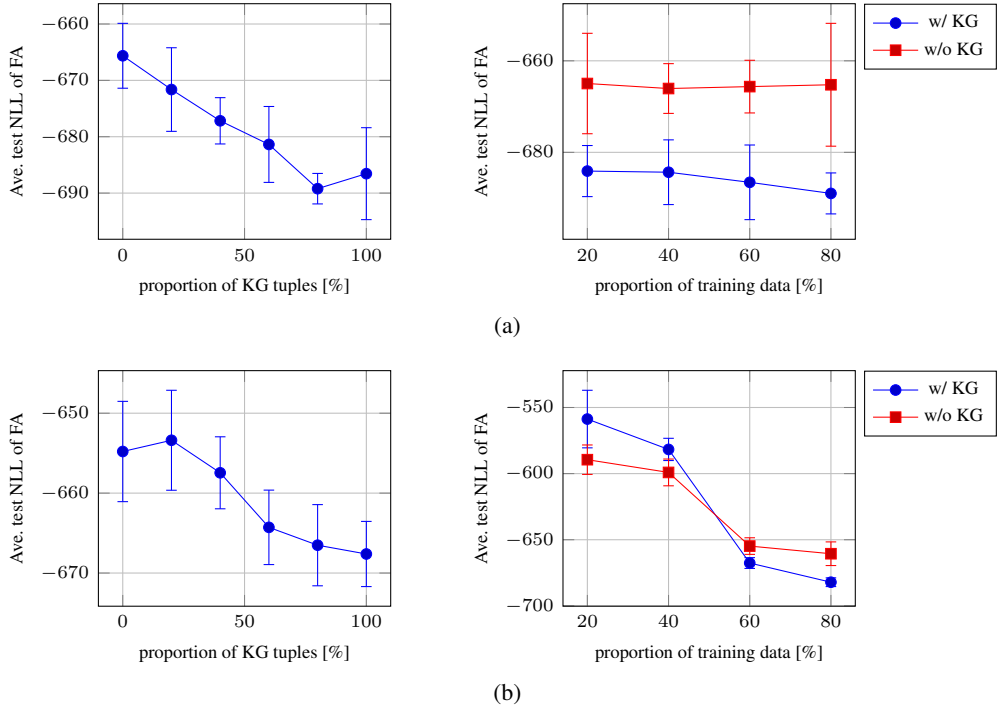

\subsection{Factor Analysis Regularized by Knowledge}

Among the probabilistic models for continuous data, the factor analysis (FA) and its special version, principal component analysis (PCA), have been favored in many applications.
As an instance of the proposed framework, we present a simple method for incorporating knowledge graphs to regularization in the parameter estimation of FA.

The observation probabilistic model of FA is often expressed as
\begin{multline}
    p(\bm{y}_1,\dots,\bm{y}_n \mid \bm{x}_1,\dots,\bm{x}_n, \theta)\\
    = \prod_{i=1}^m \prod_{j=1}^n \mathcal{N}(y_{i,j}\mid\bm{w}_i^\top\bm{x}_j + \mu_i, \ \sigma^2_i),
    \label{eq:fa:lik}
\end{multline}
where $\bm{w}_i^\top \in \mathbb{R}^{1 \times d_x}$ is the $i$-th row of the factor loading matrix of FA, $\mu_i$ is the observation mean, and $\sigma^2_i \in \mathbb{R}^+$ is the observation noise variance of the $i$-th attribute.
Moreover, $\bm{x}_j\in\mathbb{R}^{d_x}$ is the latent factor corresponding to the $j$-th object, whose prior distribution is usually set by the standard normal distribution, i.e., $\mathcal{N}(\bm{0},\bm{I})$.

Let us impose the knowledge-based prior distributions on (part of) the rows of the factor loading matrix, $\bm{w}_1,\dots,\bm{w}_{m'}$ ($m' \leq m$).
Here, instead of the normal distribution like Eq.~\eqref{eq:paramprior}, we specify the prior distribution of $\bm{w}_i$ given embedding $\bm{e}_i$ by the point mass, i.e.,
\begin{equation}
    p(\bm{w}_i \mid \bm{e}_i) = \delta (\bm{w}_i - \bm{c}_\xi(\bm{e}_i)),
    \quad\text{for}\quad i=1,\dots,m',
\end{equation}
where $\delta$ denotes the Dirac delta function. This simplification enables us to maximize the (regularized) objective just by replacing $\bm{w}_i$ with $\bm{c}_\xi(\bm{e}_i)$. We observed no significant degradation of performance due to this simplification.

In summary, an optimization problem to be solved for learning FA distantly regularized based on the knowledge graph is the maximization of
\begin{equation}
    \begin{aligned}
        &f(\bm{e}_{1:E}, M_{1:R}, \bm{w}_{1:m}, \mu_{1:m}, \sigma^2_{1:m} , \xi)\\
        &\quad = \frac1n \sum_{j=1}^n \log \mathcal{N}(\bm{y}_j \mid \bm\mu, \ \bm\Sigma)\\
        &\qquad + \frac1{\ell} \sum_{k=1}^\ell \log \Phi(\psi(\bm{e}_{\mathrm{h}_k}, \ \bm{e}_{\mathrm{t}_k}; \ M_{\mathrm{r}_k}))\\
        &\qquad + \frac1{\ell'} \sum_{k'=1}^{\ell'} \log \left( 1 - \Phi(\psi(\bm{e}_{\mathrm{h}_{k'}}, \ \bm{e}_{\mathrm{t}_{k'}}; \ M_{\mathrm{r}_{k'}})) \right),\\
        &\text{s.t.} \quad \bm{w}_i=\bm{c}_\xi(\bm{e}_i) \quad \text{for} \quad i=1,\dots,m',
    \end{aligned}
    \label{eq:final}
\end{equation}
where $\bm\mu=[\mu_1 ~ \cdots ~ \mu_m]^\top$, $\bm\Sigma=\bm{W}\bm{W}^\top+\operatorname{diag}\{\sigma^2_1,\dots,\sigma^2_m\}$, and $\bm{W}=[\bm{w}_1 ~ \cdots ~ \bm{w}_m]^\top$.
In Eq.~\eqref{eq:final}, $E$ denotes the total number of entities (including the ones corresponding to the data attributes), $R$ denotes the number of relations, and $\ell$ and $\ell'$ denote the numbers of positive and negative tuples, respectively.
The first term of Eq.~\eqref{eq:final} is obtained by marginalizing out $\bm{x}$ in Eq.~\eqref{eq:fa:lik} with the prior $\mathcal{N}(\bm{0},\bm{I})$.

The negative tuples, needed for computing Eq.~\eqref{eq:final}, are usually not available originally.
They are often created by randomly permuting the entities and/or the relations in the positive tuples (see, e.g., \cite{Nickel15,Wang17,Cai18}). In this work, we created some negative tuples per a positive tuple by replacing either the head or the tail by any other entity that exists in the knowledge graph. When permuting the entity, we chose an attribute-corresponding entity for an attribute-corresponding entity, and a non-attribute entity for a non-attribute entity, respectively.

In Eq.~\eqref{eq:final}, $\psi$ is an arbitrary score function of knowledge graph embedding learning. In the experiments introduced below, we used the simple score function of DistMult \cite{YangB15}, defined in Eq.~\eqref{eq:distmult}.
Moreover, as $\bm{c}_\xi(\cdot)$, we used an affine model
\begin{equation}
    \bm{c}_\xi(\bm{e}_i) = \bm{A}\bm{e}_i+\bm{b}
\end{equation}
with parameters $\bm{A}\in\mathbb{R}^{d_x \times d_e}$ and $\bm{b}\in\mathbb{R}^{d_x}$ used commonly for $i=1,\dots,m'$, i.e., $\xi=\{\bm{A},\bm{b}\}$.


\section{Preliminary Experiments}
\label{exp}

In this section, we provide the results of preliminary experiments to investigate how the performance of the distantly regularized models is improved.

\subsubsection{Dataset}

We used real-world data of rainfall in the world. The dataset we prepared comprises the monthly means of historical measurements of precipitation from January 1901 to December 2015 in 227 countries and regions in the world. Therefore, $n=1,380$ and $m=227$.
The dataset was created using the resource available online at Climate Change Knowledge Portal.\footnote{\url{sdwebx.worldbank.org/climateportal/}\\(retrieved 29 April 2018)}

\subsubsection{Knowledge Graph}

In order to precisely model the relationship of precipitation among countries and regions, we need meteorological knowledge to some extent. However, in our scenario, such knowledge that directly gives the relations between data attributes can hardly be obtained.
To simulate this situation, in the experiments, we utilize a knowledge graph that encodes simple geographic relations between the countries and regions.

We built a knowledge graph using the information available online,\footnote{\url{github.com/mledoze/countries/}\\(retrieved 29 April 2018)} which includes the geographic relationship between countries and regions in the world, following the scheme presented by \citetext{Bouchard15}.
The knowledge graph we used has 278 entities (including ones that do not correspond to the attributes of the data) and two types of relations, \emph{is-inside} and \emph{is-neighbor-of}; positive tuples include (Sweden, \emph{is-inside}, Europe) and (Norway, \emph{is-neighbor-of}, Sweden), for example.
And, the knowledge graph has 1,167 positive tuples comprising the entities and the relations above.

Note that, in this knowledge graph, some entities of countries like Japan have no \emph{is-neighbor-of} relations with any other entities because the \emph{is-neighbor-of} relation does not imply the cross-sea adjacency.
Therefore, if we build a single-relational graph only with the \emph{is-neighbor-of} relation, which might be utilized for methods like graph Laplacian regularization, it does not provide sufficient information on the countries that have no neighboring countries.

\subsubsection{Setups}

As for the factor analysis model distantly regularized by knowledge, the hyperparameters to be set manually are the dimensionality of latent factors, $d_x$, as well as the dimensionality of entity embeddings, $d_e$. For a fair comparison, we set the same values for them, i.e., $d_x=d_e$. And, in what follows, we show the results with $d_x=5$, which was decided empirically without any intensive search because the purpose of this experiment was just to compare the performances of a factor analysis model with and without the knowledge-based regularization.

We partitioned the dataset into the training/validation set and the test set at the rates specified below. Within the training/validation set, we always used randomly chosen 80\% for training and the remaining 20\% for validation.
When partitioning the original dataset into the training/validation set and the test set, we followed two different scenarios. One is the \texttt{Random} data-partition, in which random permutation of the 1,380 objects in each dataset was performed before the partition. Another is the \texttt{Shift} data-partition, in which no random permutation is done before the partition, and consequently, something like the covariate shift may occur, which makes generalization difficult.

We also controlled the amount of information possibly provided by the knowledge graph by changing the proportion of knowledge graph's (positive) tuples considered in the optimization of Eq.~\eqref{eq:final}. For example, when we used 80\% of the tuples, we randomly disposed 20\% of the 1,167 positive tuples and performed the optimization with the negative sampling. As the ``no knowledge graph'' limit, we ran the optimization with 0\% of the tuples.

The optimization was performed using gradient descent whose learning rate was adjusted by Adam. For the optimization, we randomly created two negative tuples per a positive tuple.
The validation set was utilized for early stopping of optimization, in which the optimization was terminated if no improvement of the loss with regard to FA (i.e., the negative of the first term of Eq.~\eqref{eq:final}) on the validation set was observed for 50 epochs, and the model that achieved the best loss was saved.
We randomly initialized the factor loading matrix of FA, the embeddings of entities, and the relation parameters. No pretraining for the embeddings was performed.


\subsubsection{Results}

In what follows, we report the average negative log likelihood (NLL) of FA for the test dataset, i.e.,
\begin{equation*}
    - \frac1{n_\text{test}} \sum_{j=1}^{n_\text{test}} \log \mathcal{N}(y_{j} \mid \bm\mu, \ \bm\Sigma),
\end{equation*}
where $n_\text{test}$ denotes the number of data points in the test set. Note that lower values of NLL are preferred.

In Figures~\ref{fig:random} and \ref{fig:shift}, we show the results under the \texttt{Random} and \texttt{Shift} data-partition scenarios, respectively. In the left plots, the test NLLs for different numbers of knowledge graph tuples used by the proposed method are shown. Basically, the better test NLLs are achieved with the larger amount of knowledge graph's information provided.
In the right plots, the test NLLs with different amounts of training data are shown. While the test NLLs are always improved by the proposed method under the \texttt{Random} scenario, no improvement is observed with somewhat small amount of training data under the \texttt{Shift} scenario.


\section{Conclusion}
\label{conclusion}

In this paper, we have proposed the \emph{knowledge-based distant regularization} framework, a general methodology to incorporate the possibly distant domain knowledge encoded in a form of knowledge graphs. In particular, we suggested the factor analysis with the knowledge-based regularization. We provided the preliminary experimental results on the improvement of the generalization capability of the regularized factor analysis.

Indeed, the proposed framework is in its infancy. There are plenty of challenges to be tackled.
For example, we need to develop a unified method for efficiently learning the models that are regularized by knowledge graphs. Also, how the generalization capability of the regularized models is improved should be analyzed theoretically and empirically. Moreover, it would be interesting to investigate how the knowledge graph embeddings are modified due to the data fed into the probabilistic models.


{\fontsize{10}{10.8}\selectfont

}


\begin{thebibliography}{}

\bibitem[\protect\citeauthoryear{Andrzejewski \bgroup et al\mbox.\egroup
  }{2011}]{Andrzejewski11}
Andrzejewski, D.; Zhu, X.; Craven, M.; and Recht, B.
\newblock 2011.
\newblock A framework for incorporating general domain knowledge into latent
  {D}irichlet allocation using first-order logic.
\newblock In {\em Proceedings of the 22nd International Joint Conference on
  Artificial Intelligence},  1171--1177.

\bibitem[\protect\citeauthoryear{Bach \bgroup et al\mbox.\egroup
  }{2017}]{Bach17}
Bach, S.~H.; He, B.; Ratner, A.; and R\'e, C.
\newblock 2017.
\newblock Learning the structure of generative models without labeled data.
\newblock In {\em Proceedings of the 34th International Conference on Machine
  Learning},  273--282.

\bibitem[\protect\citeauthoryear{Bouchard, Singh, and
  Trouillon}{2015}]{Bouchard15}
Bouchard, G.; Singh, S.; and Trouillon, T.
\newblock 2015.
\newblock On approximate reasoning capabilities of low-rank vector spaces.
\newblock In {\em Proceedings of the AAAI Spring Syposium on Knowledge
  Representation and Reasoning}.

\bibitem[\protect\citeauthoryear{Cai, Zheng, and Chang}{2018}]{Cai18}
Cai, H.; Zheng, V.~W.; and Chang, K.
\newblock 2018.
\newblock A comprehensive survey of graph embedding: Problems, techniques and
  applications.
\newblock {\em IEEE Transactions on Knowledge and Data Engineering}.
\newblock in press, arXiv:1709.07604.

\bibitem[\protect\citeauthoryear{{De Raedt}, Kersting, and
  Natarajan}{2016}]{DeRaedtBook}
{De Raedt}, L.; Kersting, K.; and Natarajan, S.
\newblock 2016.
\newblock {\em Statistical Relational Artificial Intelligence: Logic,
  Probability, and Computation}.
\newblock Morgan \& Claypool Publishers.

\bibitem[\protect\citeauthoryear{Fan \bgroup et al\mbox.\egroup }{2017}]{Fan17}
Fan, M.; Zhou, Q.; Zheng, T.~F.; and Grishman, R.
\newblock 2017.
\newblock Distributed representation learning for knowledge graphs with entity
  descriptions.
\newblock {\em Pattern Recognition Letters} 93:31--37.

\bibitem[\protect\citeauthoryear{Foulds, Kumar, and Getoor}{2015}]{Foulds15}
Foulds, J.; Kumar, S.~H.; and Getoor, L.
\newblock 2015.
\newblock Latent topic networks: A versatile probabilistic programming
  framework for topic models.
\newblock In {\em Proceedings of the 32nd International Conference on Machine
  Learning},  777--786.

\bibitem[\protect\citeauthoryear{Fung, Mangasarian, and Shavlik}{2003}]{Fung03}
Fung, G.~M.; Mangasarian, O.~L.; and Shavlik, J.~W.
\newblock 2003.
\newblock Knowledge-based support vector machine classifiers.
\newblock In {\em Advances in Neural Information Processing Systems},
  volume~15,  537--544.

\bibitem[\protect\citeauthoryear{Hu \bgroup et al\mbox.\egroup }{2016}]{Hu16}
Hu, Z.; Luo, G.; Sachan, M.; Xing, E.; and Nie, Z.
\newblock 2016.
\newblock Grounding topic models with knowledge bases.
\newblock In {\em Proceedings of the 25th International Joint Conference on
  Artificial Intelligence},  1578--1584.

\bibitem[\protect\citeauthoryear{Huang, Zhang, and Mataxas}{2011}]{Huang11}
Huang, J.; Zhang, T.; and Mataxas, D.
\newblock 2011.
\newblock Learning with structured sparsity.
\newblock {\em Journal of Machine Learning Research} 12:3371--3412.

\bibitem[\protect\citeauthoryear{Jacob, Obozinski, and Vert}{2009}]{Jacob09}
Jacob, L.; Obozinski, G.; and Vert, J.-P.
\newblock 2009.
\newblock Group lasso with overlap and graph lasso.
\newblock In {\em Proceedings of the 26th International Conference on Machine
  Learning},  433--440.

\bibitem[\protect\citeauthoryear{Kristiadi \bgroup et al\mbox.\egroup
  }{2018}]{Kristiadi18}
Kristiadi, A.; Khan, M.~A.; Lukovnikov, D.; Lehmann, J.; and Fischer, A.
\newblock 2018.
\newblock Incorporating literals into knowledge graph embeddings.
\newblock arXiv:1802.00934.

\bibitem[\protect\citeauthoryear{Mairal and Yu}{2013}]{Mairal13}
Mairal, J., and Yu, B.
\newblock 2013.
\newblock Supervised feature selection in graphs with path coding penalties and
  network flows.
\newblock {\em Journal of Machine Learning Research} 14(1):2449--2485.

\bibitem[\protect\citeauthoryear{Mei, Zhu, and Zhu}{2014}]{Mei14}
Mei, S.; Zhu, J.; and Zhu, X.
\newblock 2014.
\newblock Robust {R}eg{B}ayes: Selectively incorporating first-order logic
  domain knowledge into {B}ayesian models.
\newblock In {\em Proceedings of the 31st International Conference on Machine
  Learning}, number~1,  253--261.

\bibitem[\protect\citeauthoryear{Nickel \bgroup et al\mbox.\egroup
  }{2015}]{Nickel15}
Nickel, M.; Murphy, K.; Tresp, V.; and Gabrilovich, E.
\newblock 2015.
\newblock A review of relational machine learning for knowledge graphs.
\newblock {\em Proceedings of the IEEE} 104(1):11--33.

\bibitem[\protect\citeauthoryear{{O\~noro-Rubio} \bgroup et al\mbox.\egroup
  }{2017}]{OnoroRubio17}
{O\~noro-Rubio}, D.; Niepert, M.; Garc\'ia-Dur\'an, A.; Gonz\'alez, R.; and
  L\'opez-Sastre, R.~J.
\newblock 2017.
\newblock Representation learning for visual-relational knowledge graphs.
\newblock arXiv:1709.02314.

\bibitem[\protect\citeauthoryear{Pei, Chakraborty, and Sycara}{2015}]{Pei15}
Pei, Y.; Chakraborty, N.; and Sycara, K.
\newblock 2015.
\newblock Nonnegative matrix tri-factorization with graph regularization for
  community detection in social networks.
\newblock In {\em Proceedings of the 24th International Joint Conference on
  Artificial Intelligence},  2083--2089.

\bibitem[\protect\citeauthoryear{Pezeshkpour, Chen, and
  Singh}{2017}]{Pezeshkpour17}
Pezeshkpour, P.; Chen, L.; and Singh, S.
\newblock 2017.
\newblock Embedding multimodal relational data.
\newblock Presented in the 6th Workshop on Automated Knowledge Base
  Construction.

\bibitem[\protect\citeauthoryear{Ratner \bgroup et al\mbox.\egroup
  }{2016}]{Ratner16}
Ratner, A.; Sa, C.~D.; Wu, S.; Selsam, D.; and R\'e, C.
\newblock 2016.
\newblock Data programming: Creating large training sets, quickly.
\newblock In {\em Advances in Neural Information Processing Systems},
  volume~29,  3567--3575.

\bibitem[\protect\citeauthoryear{Richardson and Domingos}{2006}]{Richardson06}
Richardson, M., and Domingos, P.
\newblock 2006.
\newblock {M}arkov logic networks.
\newblock {\em Machine Learning} 62(1-2):107--136.

\bibitem[\protect\citeauthoryear{Schiegg, Neumann, and
  Kersting}{2012}]{Schiegg12}
Schiegg, M.; Neumann, M.; and Kersting, K.
\newblock 2012.
\newblock {M}arkov logic mixtures of {G}aussian processes: Towards machines
  reading regression data.
\newblock In {\em Proceedings of the 15th International Conference on
  Artificial Intelligence and Statistics},  1002--1011.

\bibitem[\protect\citeauthoryear{Shahid \bgroup et al\mbox.\egroup
  }{2016}]{Shahid16}
Shahid, N.; Perraudin, N.; Kalofolias, V.; Puy, G.; and Vandergheynst, P.
\newblock 2016.
\newblock Fast robust {PCA} on graphs.
\newblock {\em IEEE Journal of Selected Topics in Signal Processing}
  10(4):740--756.

\bibitem[\protect\citeauthoryear{Shang, Jiao, and Wang}{2012}]{Shang12}
Shang, F.; Jiao, L.~C.; and Wang, F.
\newblock 2012.
\newblock Graph dual regularization non-negative matrix factorization for
  co-clustering.
\newblock {\em Pattern Recognition} 45(6):2237--2250.

\bibitem[\protect\citeauthoryear{Thoma, Rettinger, and Both}{2017}]{Thoma17}
Thoma, S.; Rettinger, A.; and Both, F.
\newblock 2017.
\newblock Towards holistic concept representations: Embedding relational
  knowledge, visual attributes, and distributional word semantics.
\newblock In {\em Lecture Notes in Computer Science}, volume 10587,  694--710.

\bibitem[\protect\citeauthoryear{Tipping and Bishop}{1999}]{Tipping99}
Tipping, M.~E., and Bishop, C.~M.
\newblock 1999.
\newblock Probabilistic principal component analysis.
\newblock {\em Journal of the Royal Statistical Society: Series B}
  61(3):611--622.

\bibitem[\protect\citeauthoryear{Towell and Shavlik}{1994}]{Towell94}
Towell, G.~G., and Shavlik, J.~W.
\newblock 1994.
\newblock Knowledge-based artificial neural networks.
\newblock {\em Artificial Intelligence} 70(1-2):119--165.

\bibitem[\protect\citeauthoryear{Varma \bgroup et al\mbox.\egroup
  }{2017}]{Varma17}
Varma, P.; He, B.; Bajaj, P.; Khandwala, N.; Banerjee, I.; Rubin, D.; and R\'e,
  C.
\newblock 2017.
\newblock Generative model structure with static analysis.
\newblock In {\em Advances in Neural Information Processing Systems},
  volume~30,  239--249.

\bibitem[\protect\citeauthoryear{Varol \bgroup et al\mbox.\egroup
  }{2012}]{Varol12}
Varol, A.; Salzmann, M.; Fua, P.; and Urtasun, R.
\newblock 2012.
\newblock A constrained latent variable model.
\newblock In {\em Proceedings of the 2012 IEEE Conference on Computer Vision
  and Pattern Recognition},  2248--2255.

\bibitem[\protect\citeauthoryear{Wang \bgroup et al\mbox.\egroup
  }{2017}]{Wang17}
Wang, Q.; Mao, Z.; Wang, B.; and Guo, L.
\newblock 2017.
\newblock Knowledge graph embedding: A survey of approaches and applications.
\newblock {\em IEEE Transactions on Knowledge and Data Engineering}
  29(12):2724--2743.

\bibitem[\protect\citeauthoryear{Xiao \bgroup et al\mbox.\egroup
  }{2017}]{Xiao17}
Xiao, H.; Huang, M.; Meng, L.; and Zhu, X.
\newblock 2017.
\newblock {SSP}: Semantic space projection for knowledge graph embedding with
  text descriptions.
\newblock In {\em Proceedings of the 31st AAAI Conference on Artificial
  Intelligence},  3104--3110.

\bibitem[\protect\citeauthoryear{Xie \bgroup et al\mbox.\egroup }{2016}]{Xie16}
Xie, R.; Liu, Z.; Jia, J.; Luan, H.; and Sun, M.
\newblock 2016.
\newblock Representation learning of knowledge graphs with entity descriptions.
\newblock In {\em Proceedings of the 30th AAAI Conference on Artificial
  Intelligence},  2659--2665.

\bibitem[\protect\citeauthoryear{Yang \bgroup et al\mbox.\egroup
  }{2015}]{YangB15}
Yang, B.; Yih, W.; He, X.; Gao, J.; and Deng, L.
\newblock 2015.
\newblock Embedding entities and relations for learning and inference in
  knowledge bases.
\newblock In {\em Proceedings of the 3rd International Conference on Learning
  Representations}.

\bibitem[\protect\citeauthoryear{Yao \bgroup et al\mbox.\egroup }{2015}]{Yao15}
Yao, L.; Zhang, Y.; Wei, B.; Qian, H.; and Wang, Y.
\newblock 2015.
\newblock Incorporating probabilistic knowledge into topic models.
\newblock In {\em Advances in Knowledge Discovery and Data Mining}, volume 9078
  of {\em Lecture Notes in Computer Science},  586--597.

\bibitem[\protect\citeauthoryear{Yao \bgroup et al\mbox.\egroup }{2017}]{Yao17}
Yao, L.; Zhang, Y.; Wei, B.; Jin, Z.; Zhang, R.; Zhang, Y.; and Chen, Q.
\newblock 2017.
\newblock Incorporating knowledge graph embeddings into topic modeling.
\newblock In {\em Proceedings of the 31st AAAI Conference on Artificial
  Intelligence},  3119--3126.

\bibitem[\protect\citeauthoryear{Zhang \bgroup et al\mbox.\egroup
  }{2016}]{Zhang16}
Zhang, F.; Yuan, N.~J.; Lian, D.; Xie, X.; and Ma, W.-Y.
\newblock 2016.
\newblock Collaborative knowledge base embedding for recommender systems.
\newblock In {\em Proceedings of the 22nd ACM SIGKDD International Conference
  on Knowledge Discovery and Data Mining},  353--362.

\bibitem[\protect\citeauthoryear{Zhu, Chen, and Xing}{2014}]{Zhu14}
Zhu, J.; Chen, N.; and Xing, E.~P.
\newblock 2014.
\newblock {B}ayesian inference with posterior regularization and applications
  to infinite latent {SVM}s.
\newblock {\em Journal of Machine Learning Research} 15:1799--1847.

\end{thebibliography}
\end{document}